\documentclass{article} 
\usepackage{iclr2027_conference,times}

\usepackage{graphicx}
\usepackage{booktabs}
\usepackage{subcaption}
\usepackage{amsmath}
\usepackage{amssymb}
\usepackage{microtype}
\usepackage{hyperref}
\usepackage{url}
\usepackage{color}
\usepackage{tabularx}

\usepackage{amsmath,amsfonts,bm}

\def\eqref#1{equation~\ref{#1}}

\def\1{\bm{1}}

\DeclareMathAlphabet{\mathsfit}{\encodingdefault}{\sfdefault}{m}{sl}
\SetMathAlphabet{\mathsfit}{bold}{\encodingdefault}{\sfdefault}{bx}{n}

\title{Exploring Synchrony for Communication in Modular Neural Networks}

\author{
Nikolas Willems\textsuperscript{1}\thanks{Equal contribution.} \quad
Qiyang Sun\textsuperscript{1}\footnotemark[1]\,\,\thanks{Corresponding author.} \quad
Bj\"{o}rn Schuller\textsuperscript{1,2} \\[0.5ex]
\textsuperscript{1}Imperial College London \\
\textsuperscript{2}Technical University of Munich
}

\iclrfinalcopy 

\begin{document}

\maketitle

\begin{abstract}

One motivation for modular neural networks is to divide computation across components that learn distinct, reusable functions, which could then be recombined to support compositional generalisation. In practice, modular structure alone does not guarantee that such roles develop, and unrestricted communication between modules can undermine any specialisation that does emerge. This raises the question of how modular communication should be organised. The problem can be viewed as an analogue of the perceptual binding problem, where synchrony has been proposed as a mechanism for coordinating distributed representations. We introduce SyncNet, a modular architecture that separates module content from a dynamical phase state: learnt Kuramoto dynamics evolve the phases, while phase alignment determines communication through a shared channel. We evaluate SyncNet on Sort-of-CLEVR for relational reasoning and on SQOOP for systematic generalisation, where lower pair variety demands more recombination at test time. SyncNet gives the best held-out accuracy of the models compared at the three lowest pair varieties, by 5.1 percentage points at the hardest, and is competitive on Sort-of-CLEVR though below the strongest attention baselines. Test-time overrides show that performance depends on phases that evolve: fixed phases cost as much as no addressing at all, and permuting which module holds which region is nearly as damaging. A content-addressed variant trained from scratch nonetheless matches it, so the benefit appears to come from the modular decomposition rather than synchrony itself.

\end{abstract}

\section{Introduction}
\label{sec:intro}

Humans can understand complex problems by breaking them into smaller, reusable concepts and recombining those concepts in new situations. This ability, known as \textit{systematic} or \textit{compositional generalisation}, remains difficult for neural networks to reproduce reliably \citep{lake2018generalization, hupkes2020compositionality, dziri2023faith}. Neural networks have become much more capable with scale \citep{kaplan2020scaling, wei2022emergent}, but it remains unclear whether scale alone is enough. This motivates inductive biases that encourage networks to learn more compositional solutions \citep{goyal2022inductive, battaglia2018relational}.

For a network to recombine its components, those components need stable and reusable functions. This is \textit{functional modularity}: components perform restricted, reusable roles while remaining relatively independent of the rest of the system \citep{fodor1983modularity}. It cannot be imposed directly and must emerge through learning, so previous work has used \textit{structural modularity}, dividing a network into separately parameterised components in the hope that specialised functions develop \citep{schug2024discovering, shazeer2017outrageously, goyal2021recurrent, andreas2016neural}. This does not reliably work. Even on tasks with an explicitly modular generative structure, learnt modules do not necessarily align with that structure, and computation can become concentrated in only a few modules \citep{mittal2022modular}. How modules communicate also matters: when information can pass freely between sub-networks, increasing connectivity can weaken specialisation \citep{bena2025dynamics}. This leads to our central question: how should a modular system decide which components communicate, and how should information pass between them?

This problem resembles the \textit{binding problem}: how a distributed system represents several entities while preserving which features, roles, or values belong together \citep{greff2020binding}. In machine learning, this is usually studied as the problem of determining which features belong to the same object \citep{greff2020binding, locatello2020object}. Modular systems face a similar problem at the level of components rather than features. We therefore describe this as a \textit{computational binding problem}: structural modularity provides separate components, but does not specify how those components should coordinate. Mechanisms proposed for binding may therefore also provide mechanisms for modular communication.

One possible mechanism is synchrony. In neuroscience, synchrony has been proposed as a way of binding features belonging to the same perceptual object \citep{malsburg1981correlation, singer1995visual}. The \textit{communication through coherence} hypothesis further proposes that the relative phase of neural populations can regulate how effectively they exchange information \citep{fries2005mechanism, fries2015rhythms}, although both ideas remain debated \citep{shadlen1999synchrony, thiele2003neuronal}. In machine learning, phase variables that carry no content themselves have been used to group features into objects \citep{lowe2022complex, lowe2023rotating}, while oscillatory dynamics have also been used directly as part of a model's computation \citep{miyato2025artificial, darlow2025continuous}. In these approaches, however, synchrony acts between individual units. Whether it can instead control communication between separately parameterised modules remains, to our knowledge, unexplored.

We hypothesise that synchrony can provide this coordination and that restricting communication in this way can support compositional generalisation. Structural modularity provides components in which specialised functions may emerge, while synchrony determines how information moves between them. The key question is whether phase synchrony itself is important, or whether a similarly restrictive routing mechanism would work equally well. We make four contributions:

\begin{enumerate}
    \item We frame modular communication as a computational binding problem and propose synchrony as one possible solution.
 
    \item We introduce SyncNet, which separates module content from phase and uses phase alignment to control communication through a shared channel.
 
    \item We show that SyncNet performs competitive relational reasoning on Sort-of-CLEVR and achieves the highest held-out accuracy among the compared models at the three lowest SQOOP pair-variety settings.
 
    \item Through ablations and overrides, we show that SyncNet relies on its learnt phase dynamics, but that synchrony is not uniquely necessary.
\end{enumerate}

\section{Related work}
\label{sec:related_work}

\subsection{Modular communication}
\label{sec:modularity}

Modular architectures mainly differ in how they route information between components and combine their outputs \citep{pfeiffer2023modular}. Mixture-of-experts models route each input to a sparse subset of experts \citep{shazeer2017outrageously, fedus2022switch}. Neural module networks build computation graphs from modules associated with operations such as attending or comparing. They generalise well to unseen combinations when given the correct layout, but less reliably when that layout must also be learnt \citep{andreas2016neural, johnson2017inferring, bahdanau2019systematic}. Recurrent independent mechanisms use several relatively independent recurrent streams that compete for activation through top-$k$ attention \citep{goyal2021recurrent}. Shared workspace models instead coordinate specialised components through a small set of shared slots, limiting how much information can be broadcast globally \citep{goyal2022coordination}.

\subsection{Synchrony in neural networks}
\label{sec:synchrony-ml}

One line of work assigns each unit a phase variable that carries no content itself, separating what the unit represents from the group it belongs to. Early work introduced phase at test time to group features \citep{reichert2014neuronal}. Complex-valued autoencoders made this grouping trainable, with magnitude representing feature presence and relative phase representing group membership \citep{lowe2022complex}. Rotating Features extended scalar phase to higher-dimensional orientation vectors \citep{lowe2023rotating}. SyncNet uses the same separation between content and phase, but applies it to modules rather than individual features. It also uses phase alignment to control communication, rather than simply to identify which features belong together.

A second line of work uses oscillatory dynamics as part of the computation itself. Artificial Kuramoto Oscillatory Neurons replace scalar activations with unit vectors that evolve according to a natural frequency, coupling between oscillators, and an input-dependent stimulus, improving unsupervised object discovery and reasoning \citep{miyato2025artificial}. Winfree-oscillator networks take a similar approach using structured Winfree dynamics \citep{wonn2026}. SyncNet uses the spherical Kuramoto update of \citet{miyato2025artificial} in its perceptual front end. However, in both of these models the oscillators are individual neurons, so synchrony does not control communication between separately parameterised modules. The Continuous Thought Machine also uses synchrony, but in a different way: computation unfolds over internal steps and synchrony is measured as pairwise correlation between neuron activation histories. This synchrony forms part of the model's representation rather than controlling communication between modules \citep{darlow2025continuous}.

\section{Method}
\label{sec:method}

SyncNet is a modular architecture in which synchrony has two roles. First, it groups the spatial feature map into modules. More importantly, it controls communication between those modules through their relative phases. The design follows two main principles. Each module maintains a content state containing its current information and a phase describing its relationship to other modules. This separates what a module represents from which modules it communicates with. All modules then write to and read from a single shared channel, with the strength of communication determined by phase alignment.

Computation occurs in four main stages. An encoder turns the image and question into a feature map. The feature map is then split into spatial cells, and those cells are assigned across $M$ modules. Each module holds a content state and a phase state. The modules then exchange information over $T$ internal steps, while the content and phase states evolve. Finally, a readout maps the state of a separate head module to the output.
\begin{align}
F &= \mathcal{E}(I, \bar q) && \text{encode} \label{eq:top-encode} \\
\{X_k, h^{(0)}_k, z^{(0)}_k\}_{k=1}^{M} &= \mathcal{B}(F) && \text{bind} \label{eq:top-bind} \\
(h^{(t)}, z^{(t)}) &= \mathcal{S}(h^{(t-1)}, z^{(t-1)}, X), \quad t=1,\ldots,T && \text{communicate} \label{eq:top-step} \\
y &= \mathcal{R}(h^{(T)}_0, \bar q) && \text{read}. \label{eq:top-read}
\end{align}

\begin{figure}[h]
\centering
\includegraphics[width=\linewidth]{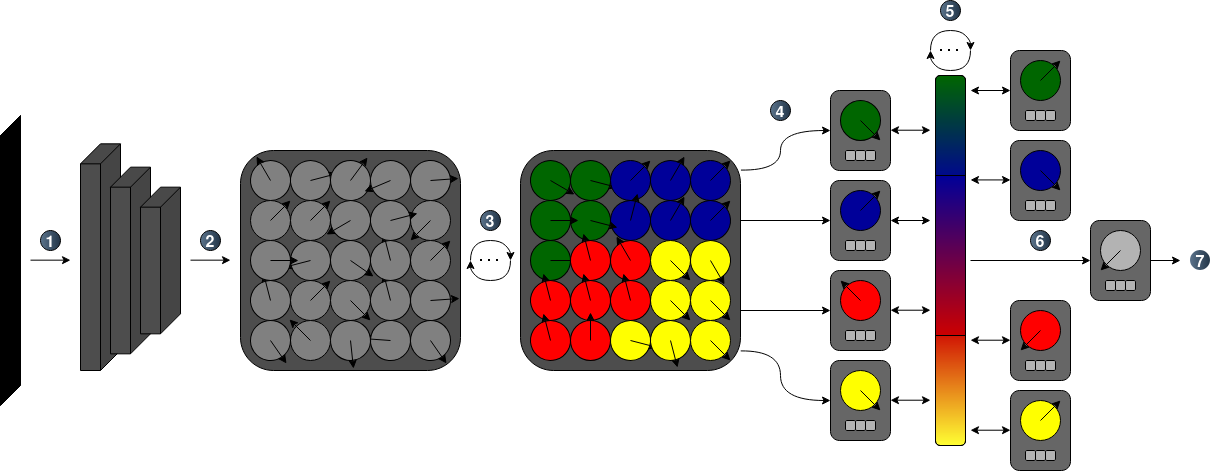}
\caption{Overview of SyncNet, shown with four modules for clarity. (1) A convolutional encoder encodes the image into a spatial feature map. (2--3) Oscillators attached to the spatial features evolve under local coupling and feature-conditioned stimuli, allowing related cells to align in phase. Colour denotes phase group throughout the figure. (4) The resulting phase groups initialise the module content states and phases. (5) Modules communicate over a shared phase-addressed channel while their phases evolve over internal steps. (6) A head module participates in the same communication process but does not observe the image directly. (7) The final answer is read from the head state.}
\label{fig:syncnet}
\end{figure}

Here, $X_k$ is the part of the feature map assigned to module $k$, $h^{(t)}_k$ its content state at internal step $t$, and $z^{(t)}_k$ its phase. Throughout, $M$ counts the perceptual modules, indexed $k=1,\ldots,M$; the head module introduced below is indexed $k=0$, so the $M+1$ participants in the shared channel are indexed $k=0,1,\ldots,M$. The remainder of this section defines each component following the numbered stages in Figure~\ref{fig:syncnet}.

\paragraph{Spherical phase.}

Phase is represented throughout as a unit vector $z_i \in S^{d-1}$ rather than as a point on a circle. Following AKOrN, it evolves using a fixed number of explicit Euler steps of the spherical Kuramoto dynamics \citep{miyato2025artificial, kuramoto1975self}:
\begin{equation}
z_i \leftarrow \Pi\!\left( z_i + \gamma \left( \Omega_i z_i + \operatorname{Proj}_{z_i} \left( \sum_j J_{ij}z_j+c_i \right) \right) \right), \label{eq:kuramoto-sphere}
\end{equation}
where $\Omega_i$ is an antisymmetric matrix controlling the oscillator's natural rotation, which keeps its contribution tangent to the sphere, $J_{ij}$ is the coupling between oscillators, $c_i$ is a conditional stimulus, and $\gamma$ is the step size. $\Omega_i$ and $J_{ij}$ are learnt in both the field and the module dynamics, with $\Omega_i$ antisymmetrised by construction; $\gamma$ is a fixed hyperparameter, with separate values for the field and the modules given in Appendix~\ref{app:models}. The tangent-space projection is $\operatorname{Proj}_{z}(v) = v-\langle v,z\rangle z$, and $\Pi(v)=v/\lVert v\rVert$ renormalises the result to unit length. Positive coupling pulls oscillators together, while negative coupling pushes them apart. At $d_x=2$ this reduces to the classical scalar Kuramoto model \citep{miyato2025artificial}.

\paragraph{(1) The encoder.}

A convolutional encoder maps the image $I$, conditioned on the encoded question $\bar q$, to a feature map $F$ over an $S\times S$ spatial grid, where $F_p$ is the feature at cell $p$. The encoder depends on the task but is shared with the baselines on that task, so differences between models do not come from different feature extractors.

\paragraph{(2--3) The oscillator field.}

Each spatial cell $p$ carries $K$ spherical oscillators of dimension $d_x$, written together as $x_p$. These are initialised from the local feature using a linear projection, $x^{(0)}_p=\Pi(W_xF_p)$ applied to each oscillator in turn, and then evolved for $T_x$ steps of Equation~\ref{eq:kuramoto-sphere} at step size $\gamma_x$. The oscillator dimension $d_x$ is a property of the field and is distinct from the module phase dimension $d$ of the module dynamics below; values for both are given in Appendix~\ref{app:models}. The update is applied across the spatial grid with learnt convolutional coupling between neighbouring cells and a feature-conditioned stimulus $C=W_cF$. This follows the oscillator layer of AKOrN \citep{miyato2025artificial}; its settings are given in Appendix~\ref{app:models}. Together, the local coupling and feature-conditioned stimulus allow related cells to move towards useful phase groupings.

\paragraph{(4) Module initialisation.}

Each module $k$ has a learnt starting phase $\bar a_k$, with a small perturbation so that its initial phase is not identical on every pass:
\begin{equation}
a^{(0)}_k = \Pi\!\left( \bar a_k+\sigma_k\epsilon_k \right), \qquad \epsilon_k\sim\mathcal{N}(0,I). \label{eq:anchor}
\end{equation}
Distinct anchors lead modules to claim different cells and so to hold different information, which is what gives them something to exchange, and learning them makes that starting point repeatable across passes. Modules then compete for spatial cells according to phase alignment over $T_b$ iterations:
\begin{equation}
\alpha^{(t)}_{kp} = \operatorname*{softmax}_{k} \left( \lambda\langle a^{(t-1)}_k,x_p\rangle+b_k(p) \right), \qquad a^{(t)}_k = \Pi\!\left( \sum_p\alpha^{(t)}_{kp}x_p \right), \label{eq:competition}
\end{equation}
where $\lambda$ controls the sharpness of the competition and is learnt from a fixed initial value, and $b_k(p)$ is a learnt spatial preference. The anchors occupy the same space as the field, so $a_k$ is likewise a set of $K$ unit vectors of dimension $d_x$, and $\langle a_k,x_p\rangle$ denotes the mean of the $K$ corresponding oscillator inner products. After the final iteration, each module receives the weighted average of its assigned features,
\begin{equation}
X_k = \frac{\sum_p\alpha^{(T_b)}_{kp}F_p}{\sum_p\alpha^{(T_b)}_{kp}}, \label{eq:module-content}
\end{equation}
with initial channel phase $z^{(0)}_k=\Pi(W_a a^{(T_b)}_k)$, where $W_a$ maps the $Kd_x$ anchor coordinates into the $d$-dimensional phase space and initial content state $h^{(0)}_k=\mathcal{I}(X_k,\bar q)$, where $\mathcal{I}$ is a two-layer multi-layer perceptron (MLP) applied to the concatenation $[X_k;\bar q]$ and is shared across modules.

\paragraph{(5) The module update.}

Module $k$ maintains a content state $h^{(t)}_k\in\mathbb{R}^{d_h}$ and a phase $z^{(t)}_k\in S^{d-1}$. Each of the $T$ internal steps consists of communication, a content update, and a phase update:
\begin{align}
r^{(t)}_k &= \mathcal{C}(m^{(t)},z^{(t-1)})_k, \qquad m^{(t)}_k=W_mh^{(t-1)}_k && \text{communicate} \label{eq:step-communicate} \\
h^{(t)}_k &= \operatorname{GRU}_k \left( [X_k;r^{(t)}_k], h^{(t-1)}_k \right) && \text{update content} \label{eq:step-update} \\
z^{(t)}_k &= \Phi(z^{(t-1)},h^{(t)})_k && \text{update phase}, \label{eq:step-phase}
\end{align}
where $m^{(t)}_k$ is module $k$'s message and $r^{(t)}_k$ is the information it receives from the shared channel $\mathcal{C}$. Each module has its own gated recurrent unit (GRU) cell $\operatorname{GRU}_k$, allowing different modules to learn different computations. These separate recurrent parameters provide SyncNet's structural modularity. The message projection $W_m$, channel, and phase dynamics are shared across modules.

\paragraph{The shared channel.}

Module $j$ writes its message in the direction of its phase, $B = \sum_j m_j z_j^\top \in \mathbb{R}^{d_m\times d}$, and module $k$ reads back the component aligned with its own phase:
\begin{equation}
\mathcal{C}(m,z)_k = \frac{1}{M}\left(Bz_k - m_k\right) = \frac{1}{M}\sum_{j\neq k} m_j\,\langle z_j, z_k\rangle, \label{eq:channel}
\end{equation}
so a sender with the same phase contributes fully, a sender in antiphase contributes negatively, and an orthogonal sender does not contribute. Subtracting $m_k$ removes the receiver's own message. All indices run over $j \in \{0,1,\ldots,M\}$: the $M$ perceptual modules together with the head module described below, which writes to and reads from the channel on the same terms. Each receiver therefore averages over the $M$ other participants, which is the normalising constant. This follows the communication-through-coherence hypothesis, where relative phase controls communication strength \citep{fries2005mechanism}.

Messages from modules with similar phases therefore combine, while messages in different phase directions remain separated. The phase dimension $d$ limits how many distinct directions can coexist. The channel is intentionally lossy, since each module receives only information aligned with its current phase, and unrestricted communication can weaken specialisation \citep{bena2025dynamics}. Equation~\ref{eq:channel} can also be viewed as a form of linear attention \citep{katharopoulos2020transformers, schlag2021linear}, with the same phase acting as both query and key, so alignment is symmetric and phases persist as dynamical states rather than being recomputed from the current content.





\paragraph{The phase dynamics.}

Module phases evolve using Equation~\ref{eq:kuramoto-sphere}, with the coupling between modules modulated by their current content states:
\begin{equation}
\Phi(z,h)_k = \Pi\!\left( z_k + \gamma \left( \Omega_k z_k + \operatorname{Proj}_{z_k} \left( \sum_j J_{kj} \kappa_{kj}(h) z_j + c_k \right) \right) \right), \label{eq:phi}
\end{equation}
\begin{equation}
\kappa_{kj}(h) = \operatorname{MLP}_\kappa\bigl([h_k;h_j]\bigr) \in (-1,1), \qquad c_k=W_ch_k. \label{eq:phi-terms}
\end{equation}
The effective coupling $J_{kj}\kappa_{kj}(h)$ is a learnt base coupling scaled by a small network that reads the two content states. Positive values pull the phases together, while negative values push them apart. The states are given in receiver--sender order, so the coupling is asymmetric, $\kappa_{kj}\neq\kappa_{jk}$. The stimulus $c_k$ also allows a module's own content to directly influence its phase. The output of $\operatorname{MLP}_\kappa$ passes through a $\tanh$, which gives the stated range.

Content therefore affects communication only indirectly. As the module states change, the couplings change, which changes the phase relationships and so the strength of communication at the next step.

\paragraph{(6--7) The head and the readout.}

The head is an additional module, indexed by $k=0$, with its own recurrent cell. It writes to and reads from the shared channel through its phase and updates that phase using the same dynamics as the other modules. However, it has no direct access to the image. Its content state is initialised from the encoded question, $h^{(0)}_0=\mathcal{I}_0(\bar q)$, its phase from a random unit vector, and its module input is fixed to $X_0=0$ because it receives no spatial cells. After $T$ steps, the prediction is read from its final state, $y = \operatorname{MLP}_y([h^{(T)}_0;\bar q])$, which is the readout $\mathcal{R}$ of Equation~\ref{eq:top-read}. Image information can therefore affect the prediction only after being assigned to a perceptual module and communicated to the head through the shared channel.

\section{Experimental setup}
\label{sec:exper_setup}

\subsection{Datasets}
\label{sec:exper_setup:datasets}

Both tasks are diagnostic probes rather than performance benchmarks, chosen because each isolates one capability under controlled conditions. Sort-of-CLEVR \citep{santoro2017simple} supplies questions as fixed binary vectors over synthetic scenes, removing language and keeping perception simple so that differences between models reflect relational computation rather than representation \citep{marza2022vision_vqa}. Its three question families, non-relational, binary and ternary, give graded relational complexity as more objects must be considered jointly; we include the ternary family of \citet{kim2018relnet}. It does not test systematic generalisation, since its questions are fixed codes that do not compose and its standard split holds out no novel combinations.

SQOOP \citep{bahdanau2019systematic} supplies that test. Each question names two of 36 objects and one of four spatial relations, and asks whether the relation holds. Difficulty is controlled by the \textit{pair variety}: the number of distinct right-hand objects paired with each left-hand object during training, with the remaining pairings held out for evaluation. Lower pair variety exposes the model to fewer combinations and demands more recombination, so the axis directly parameterises how much systematic generalisation is required. This control is why a larger benchmark would not substitute: no standard dataset exposes a comparable knob, and without it the quantity we report cannot be measured. The cost is that SQOOP tests only simple spatial relations, so strong held-out accuracy is evidence about recombination rather than about compositional reasoning in general. Example scenes and questions from both tasks are shown in Figure~\ref{fig:datasets}.

\subsection{Baselines and protocol}
\label{sec:methodology:baselines}

All models on a task share the same convolutional image encoder where possible, so that differences reflect the reasoning and communication mechanism rather than feature extraction: the encoder of \citet{santoro2017simple} on Sort-of-CLEVR and of \citet{bahdanau2019systematic} on SQOOP, with a patch embedding for the token-based models. Questions are encoded by an MLP on Sort-of-CLEVR and a long short-term memory network (LSTM) on SQOOP.

We compare against six baselines, defined in full in Appendix~\ref{app:baselines}: a \textbf{question-only} model that fixes the floor, the \textbf{convolutional} baselines CNN+MLP and Conv+LSTM \citep{santoro2017simple, bahdanau2019systematic}, the \textbf{Relation Network} \citep{santoro2017simple}, \textbf{FiLM} \citep{perez2018film}, a \textbf{Transformer} \citep{vaswani2017attention}, and the \textbf{Shared Workspace} \citep{goyal2022coordination}. The last is the closest comparison, restricting communication to a small set of persistent slots as SyncNet restricts it to one channel, but selecting by content rather than phase.

We do not report an end-to-end AKOrN baseline because AKOrN was not designed or evaluated for question-conditioned tasks \citep{miyato2025artificial}. Constructing such a baseline would therefore require introducing our own conditioning mechanism, making it difficult to distinguish the behaviour of AKOrN itself from that of our adaptation. We also do not report a neural module network. \citet{bahdanau2019systematic} show on SQOOP that one generalises well there only when both the layout and the module parametrisation are supplied by hand, and that inducing either end-to-end fails.

Optimisation settings are held fixed across models and conditions, with all deviations recorded in Appendix~\ref{app:training}; the only exceptions are the two baselines that failed to train reliably under the shared protocol and were tuned until they did. Every model is run for five seeds unless stated otherwise, and results are reported as mean $\pm$ standard deviation. For each run we select the checkpoint with the highest validation accuracy and evaluate it once on the test set. On SQOOP the validation set contains only pairings seen during training, so held-out pairings are never used for model selection.

\section{Results}
\label{sec:results}

\subsection{Compositional generalisation: SQOOP}
\label{sec:results:sqoop}

\begin{figure}[t]
\centering
\centering
\includegraphics[width=0.99\linewidth]{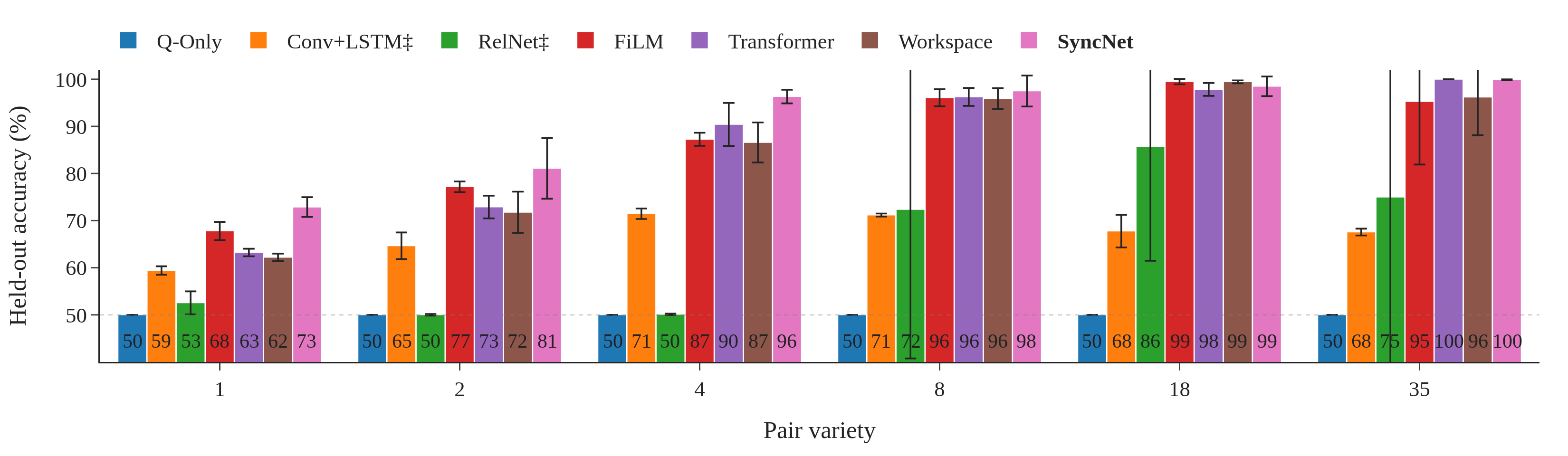}
\caption{SQOOP held-out accuracy (mean$\,\pm\,$s.d.\ over five seeds) across pair-variety settings. The Relation Network and Conv+LSTM, marked $\ddagger$, use their tuned configurations. At pair variety $35$ the held-out split matches the validation distribution.}
\label{fig:sqoop-results}
\end{figure}

Every model reaches near-perfect validation accuracy at every setting, so the comparison of interest is held-out accuracy, shown in Figure~\ref{fig:sqoop-results}. At pair varieties $8$, $18$ and $35$, FiLM, the Transformer, the Shared Workspace and SyncNet perform similarly: more pairings have been seen during training, so little recombination is required. The differences open as pair variety falls. SyncNet reaches $72.9\%$, $81.1\%$ and $96.3\%$ accuracy at pair varieties $1$, $2$ and $4$, which is $5.1$, $3.9$ and $5.9$ percentage points above the next best model at each setting. Its advantage is therefore largest exactly where the fewest combinations have been seen. The Conv+LSTM and Relation Network baselines did not train reliably under the shared protocol and are reported in their tuned configurations; Appendix~\ref{app:training} records the failure mode and the adjustments made. SyncNet gives the strongest out-of-distribution performance overall, with its clearest margin on the splits demanding the most recombination, which suggests the architecture supplies a useful inductive bias for systematic generalisation. However, these results alone do not establish that synchrony is responsible.

\subsection{Relational reasoning: Sort-of-CLEVR}
\label{sec:results:sort-of-clevr}

\begin{table}[t]
\centering
\caption{Sort-of-CLEVR: test accuracy in per cent, overall and by question family (mean over five seeds), with parameter counts in millions. Higher accuracy is better ($\uparrow$); fewer parameters is better ($\downarrow$). Our method is shown in bold. Per-subtype results are given in Appendix~\ref{app:results}.}
\label{tab:soc-results}
\setlength{\tabcolsep}{4pt}
\renewcommand{\arraystretch}{1.1}
\begin{tabular}{@{}lccccccc@{}}
\toprule
& \textbf{\textsc{SyncNet}} & \textit{Q-Only} & \textsc{CNN+MLP} & \textsc{RelNet} & \textsc{FiLM} & \textsc{Trans.} & \textsc{Works.} \\
\cmidrule(l){2-8}
\textbf{Params} (M) $\downarrow$ & 1.03 & 0.02 & 6.11 & 0.44 & 2.41 & 0.63 & 0.58 \\
\midrule
\textbf{Overall} Acc. (\%) $\uparrow$ & 91.9 & 49.4 & 66.3 & 91.3 & 95.5 & 93.7 & 94.7 \\
\midrule
Non-relational Acc. (\%) $\uparrow$ & 99.9 & 50.7 & 71.7 & 99.9 & 100.0 & 100.0 & 100.0 \\
Binary Acc. (\%) $\uparrow$ & 97.0 & 43.6 & 73.0 & 97.2 & 98.9 & 98.3 & 98.8 \\
Ternary Acc. (\%) $\uparrow$ & 78.9 & 54.1 & 54.3 & 76.7 & 87.6 & 82.9 & 85.4 \\
\bottomrule
\end{tabular}
\end{table}

Every model except CNN+MLP and the question-only baseline is near ceiling on the non-relational questions, so Table~\ref{tab:soc-results} turns on the binary and ternary families. On binary questions SyncNet reaches $97.0\%$ accuracy, the lowest of the five relational models but within two percentage points of FiLM at $98.9\%$. Overall it reaches $91.9\%$, above the Relation Network at $91.3\%$ and below the remaining three by at most $3.6$ percentage points, at $1.03$M parameters, more than the Transformer and Shared Workspace but under half of FiLM, so size does not account for the ordering.

The gap widens on ternary questions, where SyncNet reaches $78.9\%$ accuracy against $82.9\%$, $85.4\%$ and $87.6\%$ for the Transformer, Shared Workspace, and FiLM.
SyncNet is therefore a capable relational reasoner but not the strongest, and the cost of restricting communication to a single shared channel grows as information from more objects must be combined at once. This is the boundary of the mechanism: it does not confer a general reasoning advantage.

\subsection{Component ablations}
\label{sec:results:ablations}

Each variant replaces exactly one component of the main model and is trained from scratch under the shared protocol, evaluated on Sort-of-CLEVR and on SQOOP at pair variety $1$; Appendix~\ref{app:ablations} defines them. \textit{No spatial preference} sets $b_k(p)=0$, and \textit{no structural modularity} replaces the per-module recurrent cells with a single shared cell. The three communication variants complete the combinations of address source, phase or content, with combination rule, linear or softmax. \textit{Content + softmax} also replaces the shared channel with direct attention between modules, so it moves the address source and the communication topology together and isolates neither; \textit{content + linear} is the controlled comparison for the address source.

\begin{table}[t]
\centering
\caption{Component ablations. Each row reports test accuracy in per cent (mean$\,\pm\,$s.d.\ over five seeds) for the main model with one component changed; higher is better ($\uparrow$). Our method is shown in bold. SQOOP reports held-out accuracy at pair variety $1$, where $50.0$ is chance.}
\label{tab:ablation-results}
\setlength{\tabcolsep}{5pt}
\renewcommand{\arraystretch}{1.15}
\begin{tabular}{@{}lccccc@{}}
\toprule
& \multicolumn{4}{c}{\textbf{Sort-of-CLEVR}} & \textbf{SQOOP} \\
\cmidrule(lr){2-5}\cmidrule(l){6-6}
\textbf{Ablation} & \textbf{Overall} $\uparrow$ & \textbf{Non-rel.} $\uparrow$ & \textbf{Binary} $\uparrow$ & \textbf{Ternary} $\uparrow$ & \textbf{Held-out ($1$)} $\uparrow$ \\
\midrule
\textbf{\textsc{SyncNet}} (main) & 91.9 $\pm$ 1.0 & 99.9 & 97.0 & 78.9 & 72.9 $\pm$ 2.1 \\
\midrule
no spatial preference & 90.9 $\pm$ 3.6 & 99.9 & 96.3 & 76.5 & 50.0 $\pm$ 0.1 \\
no structural modularity & 90.7 $\pm$ 0.8 & 99.9 & 96.6 & 75.6 & 67.0 $\pm$ 4.6 \\
phase + softmax & 91.6 $\pm$ 0.6 & 99.9 & 97.3 & 77.6 & 69.3 $\pm$ 3.7 \\
content + linear & 90.7 $\pm$ 0.8 & 99.9 & 96.8 & 75.5 & 70.2 $\pm$ 1.5 \\
content + softmax & 91.1 $\pm$ 0.8 & 99.9 & 97.2 & 76.2 & 72.5 $\pm$ 4.4 \\
\bottomrule
\end{tabular}
\end{table}

Changes with little effect on Sort-of-CLEVR often have a much larger effect on SQOOP. Removing the spatial preference costs one percentage point on Sort-of-CLEVR but raises seed variation from $\pm 1.0$ to $\pm 3.6$, and on SQOOP drops held-out accuracy from $72.9\%$ to chance. The failure matches the optimisation problem seen in the two unstable baselines, with gradient norms falling early and runs not recovering. On SQOOP no seed escapes chance, at $50.0 \pm 0.1$, so the failure is complete rather than intermittent, while on Sort-of-CLEVR the same ablation is survivable but less reliable. The spatial preference therefore governs whether a useful decomposition can be learnt at all, rather than the capacity of that decomposition once learnt. Removing structural modularity shows the same pattern more weakly, at $90.7\%$ and $67.0\%$. Collapsing $M$ recurrent cells into one also reduces the parameter count of that component, and the shared cell is not widened to compensate, so part of this drop may reflect capacity rather than the loss of specialisation.

The communication variants behave differently. They give $91.6\%$ and $69.3\%$ for \textit{phase + softmax}, $90.7\%$ and $70.2\%$ for \textit{content + linear}, and $91.1\%$ and $72.5\%$ for \textit{content + softmax}, against $91.9\%$ and $72.9\%$ for the main model. All four are within noise of one another at five seeds, and there is no consistent ordering, since changing both the address source and the combination rule outperforms changing either alone. The address source is not what separates these models.

\subsection{Test-time overrides}
\label{sec:results:overrides}

We apply four overrides to the trained model without retraining, so that a drop in accuracy shows what the model relies on. Each acts on the perceptual modules only, and the head's phase evolves as trained. The head's phase starts from a random vector and carries no information about the scene, so freezing it too would disable the readout and confound the result.

\textit{Freeze} holds each module's phase at its post-binding value, removing evolution but keeping the initial assignment. \textit{Zero} sets every module phase to one common vector, removing addressing but leaving the messages unchanged. The two shuffle variants permute across modules after binding. \textit{Shuffle-phase} permutes the phases alone, so each module keeps its assigned content. \textit{Shuffle-binding} permutes phase and content together, so module $k$ receives module $\pi(k)$'s binding intact. Appendix~\ref{app:ablations} defines all four.

\begin{table}[t]
\centering
\caption{Test-time overrides applied to the trained main model. Values are the change in test accuracy in percentage points relative to the unmodified model (mean$\,\pm\,$s.d.\ over five seeds); a more negative value means a more damaging override.}
\label{tab:overrides}
\setlength{\tabcolsep}{6pt}
\renewcommand{\arraystretch}{1.15}
\begin{tabular}{@{}lcccc@{}}
\toprule
\textbf{Task} & \textbf{Freeze} & \textbf{Zero} & \textbf{Shuffle-phase} & \textbf{Shuffle-binding} \\
\midrule
Sort-of-CLEVR & -42.7 $\pm$ 25.4 & -41.9 $\pm$ 25.9 & -5.6 $\pm$ 10.0 & -31.9 $\pm$ 3.7 \\
SQOOP ($1$) & -22.8 $\pm$ 2.7 & -24.2 $\pm$ 3.6 & -1.9 $\pm$ 1.8 & -23.2 $\pm$ 2.4 \\
\bottomrule
\end{tabular}
\end{table}

Freeze and Zero cost much the same, $42.7$ and $41.9$ percentage points of accuracy on Sort-of-CLEVR and $22.8$ and $24.2$ on SQOOP. Both leave SQOOP at chance and Sort-of-CLEVR at the question-only floor of $49.4\%$. Phase addressing is therefore load-bearing, but fixed phases are worth no more than no addressing at all. What the model uses is how the phases evolve, not how they differ at the first step.

Permuting the phases alone costs $5.6$ percentage points on Sort-of-CLEVR and $1.9$ on SQOOP. Each module keeps its content, and the set of phases and the angles between them are unchanged, so only which module holds which phase differs. The model is close to indifferent: it depends on how phases relate as they evolve, not on which phase a module starts from.

Permuting phase and content together costs $31.9$ percentage points and $23.2$, again leaving SQOOP at chance. The pairing of phase to content survives, so what changes is which module holds which region. The modules are therefore not interchangeable, since the private cells and the readout depend on where a region's binding arrives. This is our most direct evidence that the modules specialise, and it agrees with the shared-cell ablation above.

Taken with the ablations, this locates the mechanism. Synchrony is load-bearing once learnt, and removing phase evolution from a trained model drops it to the floor. It is not uniquely necessary, since a content-addressed model trained from scratch reaches comparable accuracy. What the two share is the modular decomposition, and disturbing that decomposition costs as much as removing synchrony. The evidence supports the stability of the decomposition rather than the mechanism that addresses it.

Appendix~\ref{app:results} reports sweeps over the phase dimension, the number of internal steps and the number of modules. Accuracy rises with $d$ on Sort-of-CLEVR and flattens once three-way interactions can be separated, SQOOP benefits from more internal steps, and the module count has little effect up to $M=8$.

\section{Conclusion}
\label{sec:conclusion}

Synchrony can coordinate communication between the components of a modular network. SyncNet performs non-trivial relational reasoning while restricting all inter-module traffic to a single phase-addressed channel, and reaches the highest held-out accuracy among the models compared at the three lowest pair-variety settings, with its clearest margin where the fewest combinations have been seen during training. Overrides confirm the trained model depends on phases that evolve: holding them fixed costs as much as removing addressing altogether, and permuting which module holds which region costs almost as much again. Ablations show that a content-addressed variant learns an equally good solution from scratch, and that a stable modular decomposition matters more than the mechanism used to address it.

Three limits bound these claims. The evidence comes from two diagnostic tasks chosen for the control they afford, and we have no evidence about behaviour at scale. Functional modularity is only partly measured. The overrides show that the modules are not interchangeable, since permuting which module holds which region costs a trained model most of its accuracy, but they say nothing about what any module computes, so whether the roles acquired are reusable across tasks is untested and is the extension we would pursue first. And while synchrony is sufficient and load-bearing once learnt, these experiments do not show it is preferable to more general routing.

\newpage

\subsection*{AI use statement}

We used generative AI assistants during this project, and disclose their use as follows. They were used to format and debug \LaTeX{} tables and figures, with every reported number checked against its source; to assist with sentence-level wording and to identify repetition, with all changes reviewed for consistency with the intended content; to explain technical concepts during the research process, with every claim in this paper verified against the primary source; to help read other research codebases when adapting baseline models, so that those adaptations remained faithful to the original implementations; and to help restructure and simplify existing code for readability without changing its behaviour. Generative AI can be fluent and wrong at the same time, so all AI-generated suggestions were independently reviewed before use. The framing of the problem, the research questions, the methodology, the experimental design, the implementation decisions, and the analysis, interpretation and conclusions presented here are the authors' own. Specifically, ChatGPT (GPT-5.6 Sol) was used in relation to the above.

\subsection*{Ethics statement}

This work raises no ethical concerns. It involves no human participants and no personal data, and both datasets are generated synthetically, so no questions of consent or privacy arise. The datasets contain geometric shapes and alphanumeric glyphs, so the representation concerns attaching to natural image corpora do not apply. The models are small relational reasoners trained on synthetic scenes and have no application we are aware of that carries a foreseeable risk of harm. The authors declare no conflicts of interest and no sponsorship that could have influenced the work. All experiments were run on a single GPU, with individual training runs averaging roughly 45 minutes, so the compute cost of the work was modest.

\subsection*{Reproducibility statement}

The architecture is specified in full in Section~\ref{sec:method}, with the oscillator field's settings given in Appendix~\ref{app:models}. Dataset generator settings for both tasks, including every pair-variety setting, are in Appendix~\ref{app:data}; both datasets are synthetic and can be regenerated deterministically. Baseline architectures are in Appendix~\ref{app:baselines}, per-model configurations in Appendix~\ref{app:models}, and the shared optimisation protocol in Appendix~\ref{app:training}, which also records every deviation from it by model and task. The ablation variants and the test-time overrides are defined exactly in Appendix~\ref{app:ablations}. Every reported number is a mean over five seeds with the standard deviation given where applicable, and checkpoint selection is described in Appendix~\ref{app:training}. The source code is included as a zipped archive in the supplementary material.

\bibliography{iclr2027_conference}
\bibliographystyle{iclr2027_conference}

\appendix

\section{Dataset configurations}
\label{app:data}
\begin{figure}[t]
\centering
    \begin{subfigure}[b]{0.65\textwidth}
        \centering
        \begin{minipage}[c]{0.21\linewidth}
            \centering
            \fbox{\includegraphics[width=\linewidth]{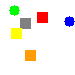}}
        \end{minipage}\hfill
        \begin{minipage}[c]{0.75\linewidth}
            \footnotesize
            \begin{tabular}{@{}l p{0.72\linewidth}@{}}
                \toprule
                \textbf{family} & \textbf{question} \\
                \midrule
                Non-rel. & Shape of the blue object? \textbf{circle} \\
                Binary & Shape of the object closest to the green object? \textbf{rectangle} \\
                Ternary & How many objects lie in the box between the blue and yellow objects? \textbf{1} \\
                \bottomrule
            \end{tabular}
        \end{minipage}
        \caption{Sort-of-CLEVR}
        \label{fig:soc-example}
    \end{subfigure}\hfill
    \begin{subfigure}[b]{0.32\textwidth}
    \centering
        \begin{minipage}[c]{0.42\linewidth}
            \centering
            \includegraphics[width=\linewidth]{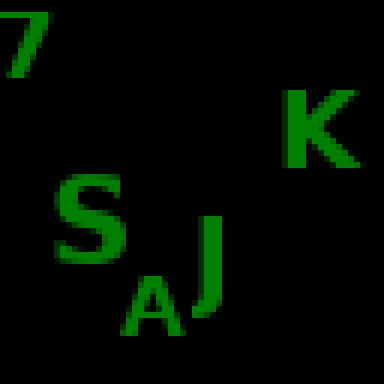}
        \end{minipage}\hfill
        \begin{minipage}[c]{0.54\linewidth}
            \footnotesize
            \begin{tabular}{@{}p{0.95\linewidth}@{}}
                \toprule
                \textbf{question} \\
                \midrule
                \textsf{7 above S}? \\
                \hspace{1em}\textbf{yes} \\
                \addlinespace
                \textsf{7 left of S}? \\
                \hspace{1em}\textbf{no} \\
                \bottomrule
            \end{tabular}
        \end{minipage}
        \caption{SQOOP}
        \label{fig:sqoop-examples}
    \end{subfigure}
    \caption{Example scenes and questions from the two tasks. Generator settings are given in Table~\ref{tab:task-config}.}
    \label{fig:datasets}
\end{figure}

Figure~\ref{fig:datasets} shows example scenes and questions from both tasks, and Table~\ref{tab:task-config} gives the generator settings used to produce them. Separate datasets are generated for each pair-variety setting of SQOOP.

\begin{table}[ht]
\centering
\begin{tabular}{lll}
\toprule
Setting & Sort-of-CLEVR & SQOOP \\
\midrule
Training examples & 1{,}080{,}000 (36{,}000 scenes) & 1{,}080{,}000 \\
Evaluation examples per split & 30{,}000 (1{,}000 scenes) & 25{,}600 \\
Image size (px) & $75 \times 75$ & $64 \times 64$ \\
Objects per scene & 6 & 5 \\
Object size (px) & 5 & 10 -- 15 \\
Questions per scene & 30 (10 per family) & --- \\
Object vocabulary & --- & 36 \\
Spatial relations & --- & 4 \\
Pair variety & --- & [1, 2, 4, 8, 18, 35] \\
\bottomrule
\end{tabular}
\caption{Dataset Generator configuration for Sort-of-CLEVR and SQOOP, separate datasets are generated for each of the listed pair-variety settings.}
\label{tab:task-config}
\end{table}

\section{Baseline architectures}
\label{app:baselines}

Notation follows Section~\ref{sec:method}: $I$ is the image, $q$ the question, $\bar q$ its encoding, $F=\operatorname{enc}(I)$ the image feature map, $F_p$ the feature at spatial location $p$, and $\xi_p$ its coordinates. Components shared across models are listed in Table~\ref{tab:shared-config} of Appendix~\ref{app:models}, which also gives the per-model configurations; deviations from the shared optimisation protocol are recorded in Appendix~\ref{app:training}.

\paragraph{Question-only.}

The question-only baseline predicts the answer from the encoded question without using the image. It shows how much performance can be obtained from the question and answer distributions alone. This is particularly useful on Sort-of-CLEVR, where answer priors differ across question types.

\paragraph{Convolutional baselines.}

The convolutional baselines test how much of each task can be solved without an explicit relational or communication mechanism. The image feature map is flattened, concatenated with the encoded question, and passed through an MLP:
\begin{equation}
\hat y = \operatorname{MLP}\left([\operatorname{vec}(F);\bar q]\right).
\label{eq:baseline-conv}
\end{equation}








On Sort-of-CLEVR, the question is encoded with an MLP. On SQOOP, it is encoded with an LSTM. These give the CNN+MLP and Conv+LSTM baselines used in the original papers \citep{santoro2017simple, bahdanau2019systematic}.

\paragraph{Relation Network.}

The Relation Network explicitly models pairwise relations between spatial features \citep{santoro2017simple}. Each ordered pair of features is processed by the same relation function, conditioned on the question:
\begin{equation}
\hat y = f_{\phi}\left(\sum_{p}\sum_{p'} g_{\theta}\left([o_p;o_{p'};\bar q]\right)\right), \qquad o_p = [F_p;\xi_p].
\label{eq:baseline-relnet}
\end{equation}

Here, $g_{\theta}$ represents the relation between each pair of spatial features, while $f_{\phi}$ maps the combined relations to the final answer. Unlike SyncNet, relational interactions are built directly into the architecture.

\paragraph{FiLM.}

FiLM uses the question to modulate intermediate visual features \citep{perez2018film}. Writing $H^{(0)}=F$, the question representation produces feature-wise scaling and shifting parameters $(\gamma^{(b)},\beta^{(b)})$ for each residual block:
\begin{align}
u^{(b)} &= \operatorname{ReLU}\left(W_1^{(b)} * [H^{(b-1)};\xi]\right), \\
H^{(b)} &= \operatorname{ReLU}\left((1+\gamma^{(b)})\odot\operatorname{BN}\left(W_2^{(b)} * u^{(b)}\right)+\beta^{(b)}\right)+u^{(b)}.
\label{eq:baseline-film}
\end{align}

Here, $\gamma^{(b)}$ and $\beta^{(b)}$ follow the notation of \citet{perez2018film} and are unrelated to the integration step size and the competition sharpness of Section~\ref{sec:method}. The question can therefore influence visual processing throughout the network. Unlike the Relation Network, FiLM does not explicitly model pairwise relations.

\paragraph{Transformer.}

The Transformer provides a general-purpose baseline based on self-attention \citep{vaswani2017attention}. The image is represented as a sequence of spatial tokens with positional information and passed through a standard Transformer encoder.

On Sort-of-CLEVR, the encoded question is added as another token. On SQOOP, the question representation is concatenated to each image token. In both cases, the final answer is predicted from a classification token.

Unlike SyncNet, the Transformer determines interactions directly from token content through self-attention. It does not maintain a separate state controlling communication.

\paragraph{Shared Workspace.}

The Shared Workspace restricts communication through a small set of persistent workspace slots \citep{goyal2022coordination}. Our implementation is based on the Transformer baseline and uses the same tokenisation, positional information, and question conditioning. However, direct self-attention between tokens is replaced by communication through the workspace.

During the write step, the workspace slots attend over the input tokens. Only a fixed fraction of tokens with the highest attention weights contribute to each slot. During the broadcast step, the tokens attend over the updated workspace:
\begin{align}
W^{(\ell+1)} &= \operatorname{Update}\left(W^{(\ell)},\operatorname{MHA}_{\text{top-}k}\left(W^{(\ell)},Z^{(\ell)}\right)\right),
\label{eq:baseline-workspace} \\
Z^{(\ell+1)} &= \operatorname{Block}\left(Z^{(\ell)}+\operatorname{MHA}\left(Z^{(\ell)},W^{(\ell+1)}\right)\right).
\end{align}

The same layer is applied recurrently with shared weights, following the recurrent form of the original architecture. This is the closest baseline to SyncNet because both restrict communication through a shared channel. The main difference is that the Shared Workspace controls access using content-based attention, while SyncNet uses phase.

\section{Model configurations}
\label{app:models}

\begin{table}[ht]
\centering
\small
\setlength{\tabcolsep}{5pt}
\renewcommand{\arraystretch}{1.15}

\begin{tabularx}{\textwidth}{
    @{}
    >{\raggedright\arraybackslash}p{0.18\textwidth}
    >{\raggedright\arraybackslash}X
    >{\raggedright\arraybackslash}X
    @{}
}
\toprule
\textbf{Component} & \textbf{Sort-of-CLEVR} & \textbf{SQOOP} \\
\midrule

\addlinespace

Shared encoder
& Four $3 \times 3$ convolutions, stride 2, 128 channels, batch norm and ReLU; $5 \times 5$ output map
& Six $3 \times 3$ convolutions, two max-pools, 64 channels, batch norm and ReLU; $16 \times 16$ output map \\

\addlinespace

Patch embedding
& Single $5 \times 5$ convolution, stride 5; $15 \times 15$ tokens
& Single $4 \times 4$ convolution, stride 4; $16 \times 16$ tokens \\

\addlinespace

Question encoder
& Identity or 2-layer MLP for CNN+MLP
& LSTM \\

\bottomrule
\end{tabularx}

\caption{Components shared across models.}
\label{tab:shared-config}
\end{table}

\begin{table}[ht]
\centering
\small
\begin{tabular}{@{}lcc@{}}
\toprule
\textbf{Setting} & \textbf{Sort-of-CLEVR} & \textbf{SQOOP} \\
\midrule
question encoder & identity & LSTM  \\
hidden dimension & 128 & 128 \\
layers & 2 & 2 \\
parameters & 20k & 118k \\
\bottomrule
\end{tabular}
\caption{Question-only model configuration for all experiments on Sort-of-CLEVR and SQOOP.}
\label{tab:cfg-qonly}
\end{table}

\begin{table}[ht]
\centering
\small
\begin{tabular}{@{}lcc@{}}
\toprule
\textbf{Setting} & \textbf{Sort-of-CLEVR (CNN+MLP)} & \textbf{SQOOP (Conv+LSTM)} \\
\midrule
encoder & shared, 128 channels & shared, 64 channels \\
question encoder & MLP & LSTM  \\
classifier hidden dimension & 256 & 256 \\
parameters & 6.11M & 9.32M \\
\bottomrule
\end{tabular}
\caption{Convolutional baseline configuration for all experiments on Sort-of-CLEVR and SQOOP.}
\label{tab:cfg-conv}
\end{table}

\begin{table}[ht]
\centering
\small
\begin{tabular}{@{}lcc@{}}
\toprule
\textbf{Setting} & \textbf{Sort-of-CLEVR} & \textbf{SQOOP} \\
\midrule
encoder & shared, 128 channels & shared, 64 channels \\
question encoder & identity & LSTM  \\
$g_\theta$ hidden dimension & 256 & 256 \\
$f_\phi$ hidden dimension & 256 & 256 \\
pairing grid (max-pooled) & $5 \times 5$ & $8 \times 8$ \\
coordinate channels & yes & yes \\
parameters & 0.44M & 1.26M \\
\bottomrule
\end{tabular}
\caption{Relation Network configuration for all experiments on Sort-of-CLEVR and SQOOP.}
\label{tab:cfg-relnet}
\end{table}

\begin{table}[ht]
\centering
\small
\begin{tabular}{@{}lcc@{}}
\toprule
\textbf{Setting} & \textbf{Sort-of-CLEVR} & \textbf{SQOOP} \\
\midrule
encoder & shared, 128 channels & shared, 64 channels \\
question encoder & identity & LSTM  \\
residual blocks & 4 & 4 \\
block channels & 128 & 128 \\
classifier channels & 512 & 512 \\
classifier hidden dimension & 1024 & 1024 \\
coordinate channels & yes & yes \\
parameters & 2.41M & 3.26M \\
\bottomrule
\end{tabular}
\caption{FiLM configuration for all experiments on Sort-of-CLEVR and SQOOP.}
\label{tab:cfg-film}
\end{table}

\begin{table}[ht]
\centering
\small
\begin{tabular}{@{}lcc@{}}
\toprule
\textbf{Setting} & \textbf{Sort-of-CLEVR} & \textbf{SQOOP} \\
\midrule
encoder & patches ($5 \times 5$) & patches ($4 \times 4$) \\
question encoder & identity & token embedding, 32-d \\
hidden dimension & 128 & 128 \\
encoder layers & 4 & 4 \\
attention heads & 4 & 4 \\
feed-forward multiplier & 2 & 4 \\
dropout & 0 & 0 \\
positional encoding & learnt 2D & learnt 2D \\
question conditioning & token & broadcast-concatenate \\
readout & \textsc{cls} & \textsc{cls} \\
shared layer weights & no & no \\
parameters & 0.63M & 1.01M \\
\bottomrule
\end{tabular}
\caption{Transformer configuration for all experiments on Sort-of-CLEVR and SQOOP.}
\label{tab:cfg-transformer}
\end{table}

\begin{table}[ht]
\centering
\small
\begin{tabular}{@{}lcc@{}}
\toprule
\textbf{Setting} & \textbf{Sort-of-CLEVR} & \textbf{SQOOP} \\
\midrule
encoder & patches ($5 \times 5$) & patches ($4 \times 4$) \\
question encoder & identity & token embedding, 32-d \\
hidden dimension & 128 & 128 \\
layers (one shared layer, applied 4 times) & 4 & 4 \\
attention heads & 4 & 4 \\
feed-forward multiplier & 2 & 2 \\
question conditioning & token & token \\
readout & \textsc{cls} & \textsc{cls} \\
workspace slots & 8 & 8 \\
top-$k$ fraction & 0.2 & 0.2 \\
key size & 32 & 32 \\
parameters & 0.58M & 0.71M \\
\bottomrule
\end{tabular}
\caption{Shared Workspace configuration for all experiments on Sort-of-CLEVR and SQOOP.}
\label{tab:cfg-workspace}
\end{table}

\begin{table}[ht]
\centering
\small
\begin{tabular}{@{}lcc@{}}
\toprule
\textbf{Setting} & \textbf{Sort-of-CLEVR} & \textbf{SQOOP} \\
\midrule
encoder & shared, $1 \times 1$ adapter to 64 & shared, 64 channels \\
question encoder & identity & token embedding, 32-d \\
modules $M$ & 6 & 6 \\
phase dimension $d$ & 6 & 6 \\
internal steps $T$ & 8 & 8 \\
step size $\gamma$ & 0.5 & 0.5 \\
module state $d_h$ & 96 & 96 \\
message width $d_m$ & 4 & 4 \\
oscillators per cell & 16, each 4-d (64 channels) & 16, each 4-d (64 channels) \\
field steps $T_x$, step size $\gamma_x$ & 8, 1.0 & 8, 1.0 \\
parameters & 1.03M & 0.83M \\
\bottomrule
\end{tabular}
\caption{SyncNet configuration for all experiments on Sort-of-CLEVR and SQOOP.}
\label{tab:canonical}
\end{table}

The oscillator field applies Equation~\ref{eq:kuramoto-sphere} across the $S\times S$ spatial grid. Each cell carries $K$ oscillators of dimension $d_x$, so the field has $Kd_x$ channels at each spatial position and each oscillator is renormalised to unit length independently. The coupling $J$ is a learnt convolution over the grid, so a cell is coupled only to its neighbourhood and the same coupling applies at every position, and the stimulus is $C=W_cF$, a pointwise linear map of the feature map, which ties each oscillator's preferred direction to the feature at its own cell. Both $\Omega$ and $J$ are learnt, as in Section~\ref{sec:method}.

\section{Training protocol}
\label{app:training}

The optimisation settings are given in Table~\ref{tab:training-config} and are held fixed across models and experimental conditions wherever possible. The only exceptions are baselines that failed to train under the shared protocol. In these cases, the optimisation settings were adjusted until training became reliable, with all deviations recorded in Table~\ref{tab:protocol-deviations}.

For each run, the checkpoint with the highest validation accuracy is selected and evaluated once on the test set. The validation set is always drawn from the same distribution as the corresponding training data. On Sort-of-CLEVR, the validation and test sets contain separate scenes sampled from the same distribution. On SQOOP, validation examples contain object pairings observed during training, while the test set contains held-out pairings that are not used for model selection.

Models are implemented in PyTorch and trained using bf16 mixed precision. Full dataset-generation settings are provided in Table~\ref{tab:task-config}, while the complete configuration of each model is given in Appendix~\ref{app:models}.

All runs used a single GPU, with an average training run taking roughly 45 minutes.

\begin{table}[ht]
\centering
\begin{tabular}{@{}ll@{}}
\toprule
\textbf{Setting} & \textbf{Value} \\
\midrule
Optimiser & AdamW \\
Learning rate & $3 \times 10^{-4}$ \\
Weight decay & 0.01 \\
Schedule & 2{,}000 warmup steps, then cosine to zero \\
Batch size & 256 \\
Gradient clipping & 1.0 \\
Optimisation steps & 100{,}000 \\
\bottomrule
\end{tabular}
\caption{The shared training configurations. Every reported result uses these settings except the rows listed in Table~\ref{tab:protocol-deviations}.}
\label{tab:training-config}
\end{table}

\begin{table}[ht]
\centering
\small
\begin{tabular}{@{}llllll@{}}
\toprule
\textbf{Model} & \textbf{Task} & \textbf{Effective batch} & \textbf{Learning rate} & \textbf{Weight decay} & \textbf{Steps} \\
\midrule
Transformer & SQOOP & --- & $10^{-4}$ & --- & --- \\
Workspace & SQOOP & --- & $10^{-4}$ & --- & --- \\
RelNet & SQOOP & 2{,}048 & --- & --- & 20{,}000 \\
Conv+LSTM & SQOOP & 2{,}048 & $10^{-4}$ & 0 & 40{,}000 \\
\bottomrule
\end{tabular}
\caption{Deviations from the shared configuration, by model and task. The dash indicates shared value.}
\label{tab:protocol-deviations}
\end{table}

\section{Additional results}
\label{app:results}

\subsection{Sort-of-CLEVR by subtype}

\begin{table}[ht]
\centering

\label{tab:soc-results-full}
\setlength{\tabcolsep}{4pt}
\renewcommand{\arraystretch}{1.1}
\begin{tabular}{@{}lccccccc@{}}
\toprule
& \textbf{\textsc{SyncNet}} & \textit{Q-Only} & \textsc{CNN+MLP} & \textsc{RelNet} & \textsc{FiLM} & \textsc{Trans.} & \textsc{Works.} \\

\midrule
\textbf{Non-relational} & 99.9 & 50.7 & 71.7 & 99.9 & 100.0 & 100.0 & 100.0 \\
\hspace{1em}Query shape & 100.0 & 49.3 & 65.0 & 100.0 & 100.0 & 100.0 & 100.0 \\
\hspace{1em}Left of centre & 99.9 & 50.6 & 75.8 & 99.9 & 100.0 & 100.0 & 100.0 \\
\hspace{1em}Top half & 99.9 & 52.1 & 74.6 & 99.9 & 100.0 & 100.0 & 100.0 \\
\midrule
\textbf{Binary} & 97.0 & 43.6 & 73.0 & 97.2 & 98.9 & 98.3 & 98.8 \\
\hspace{1em}Closest shape & 94.4 & 50.1 & 66.6 & 96.1 & 99.0 & 96.6 & 97.8 \\
\hspace{1em}Furthest shape & 96.6 & 50.4 & 74.0 & 95.5 & 97.6 & 98.5 & 98.6 \\
\hspace{1em}Count same shape & 99.9 & 29.9 & 78.7 & 100.0 & 99.9 & 100.0 & 100.0 \\
\midrule
\textbf{Ternary} & 78.9 & 54.1 & 54.3 & 76.7 & 87.6 & 82.9 & 85.4 \\
\hspace{1em}Count in box & 88.9 & 69.6 & 69.0 & 87.4 & 97.3 & 91.1 & 94.0 \\
\hspace{1em}On band & 85.8 & 51.8 & 56.1 & 83.6 & 93.7 & 87.4 & 90.5 \\
\hspace{1em}Count obtuse & 62.2 & 40.5 & 37.3 & 58.7 & 71.3 & 69.8 & 71.4 \\
\bottomrule
\end{tabular}
\end{table}

The ternary gap on Sort-of-CLEVR is concentrated in the \textit{Count obtuse} subtype, where SyncNet reaches $62.2\%$ accuracy against $69.8\%$, $71.4\%$ and $71.3\%$ for the Transformer, Shared Workspace, and FiLM. The question asks how many third objects form an obtuse triangle with two queried objects, so information from several objects must be combined simultaneously. It is the hardest subtype for every model, so task difficulty contributes alongside the architectural restriction.

\subsection{Phase capacity and communication time}
\label{sec:results:hyper_param_sweeps}

\begin{figure}[ht]
\centering
\begin{subfigure}[t]{0.33\textwidth}
    \centering
    \includegraphics[width=\linewidth]{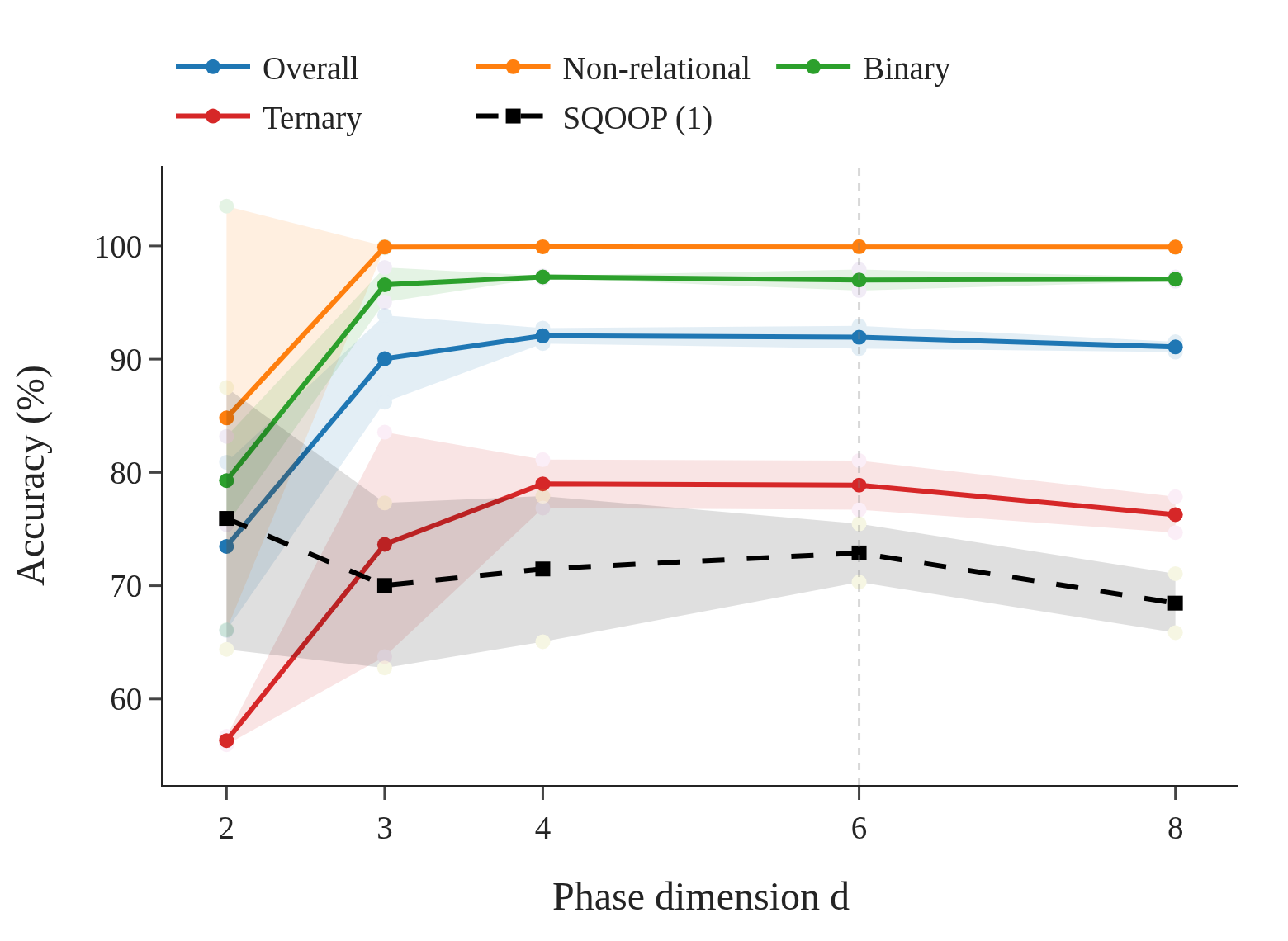}
    \caption{Phase dimension $d$.}
    \label{fig:soc-ablations-phase-dim}
\end{subfigure}\hfill
\begin{subfigure}[t]{0.33\textwidth}
    \centering
    \includegraphics[width=\linewidth]{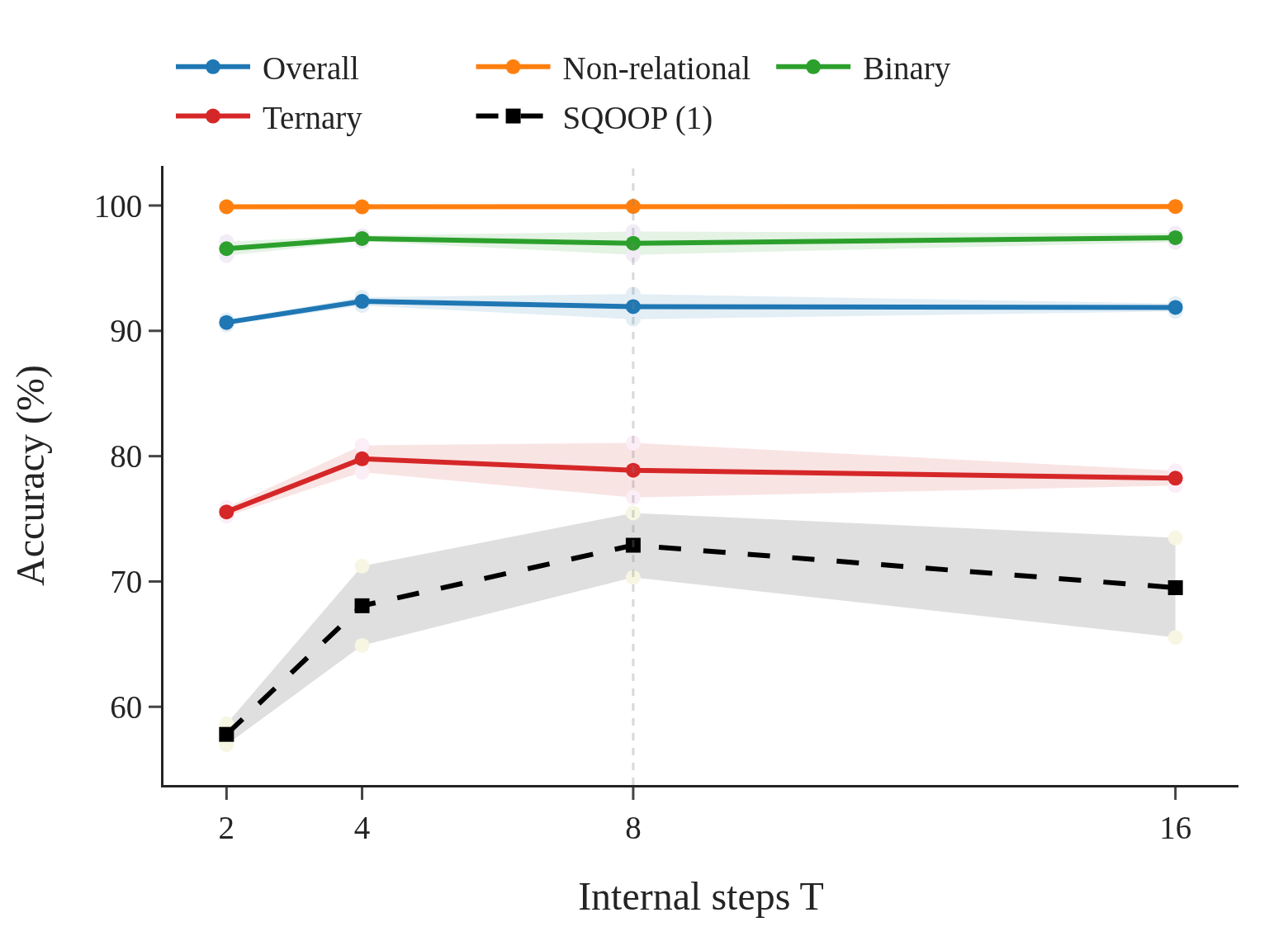}
    \caption{Internal steps $T$.}
    \label{fig:soc-ablation-t-train}
\end{subfigure}\hfill
\begin{subfigure}[t]{0.33\textwidth}
    \centering
    \includegraphics[width=\linewidth]{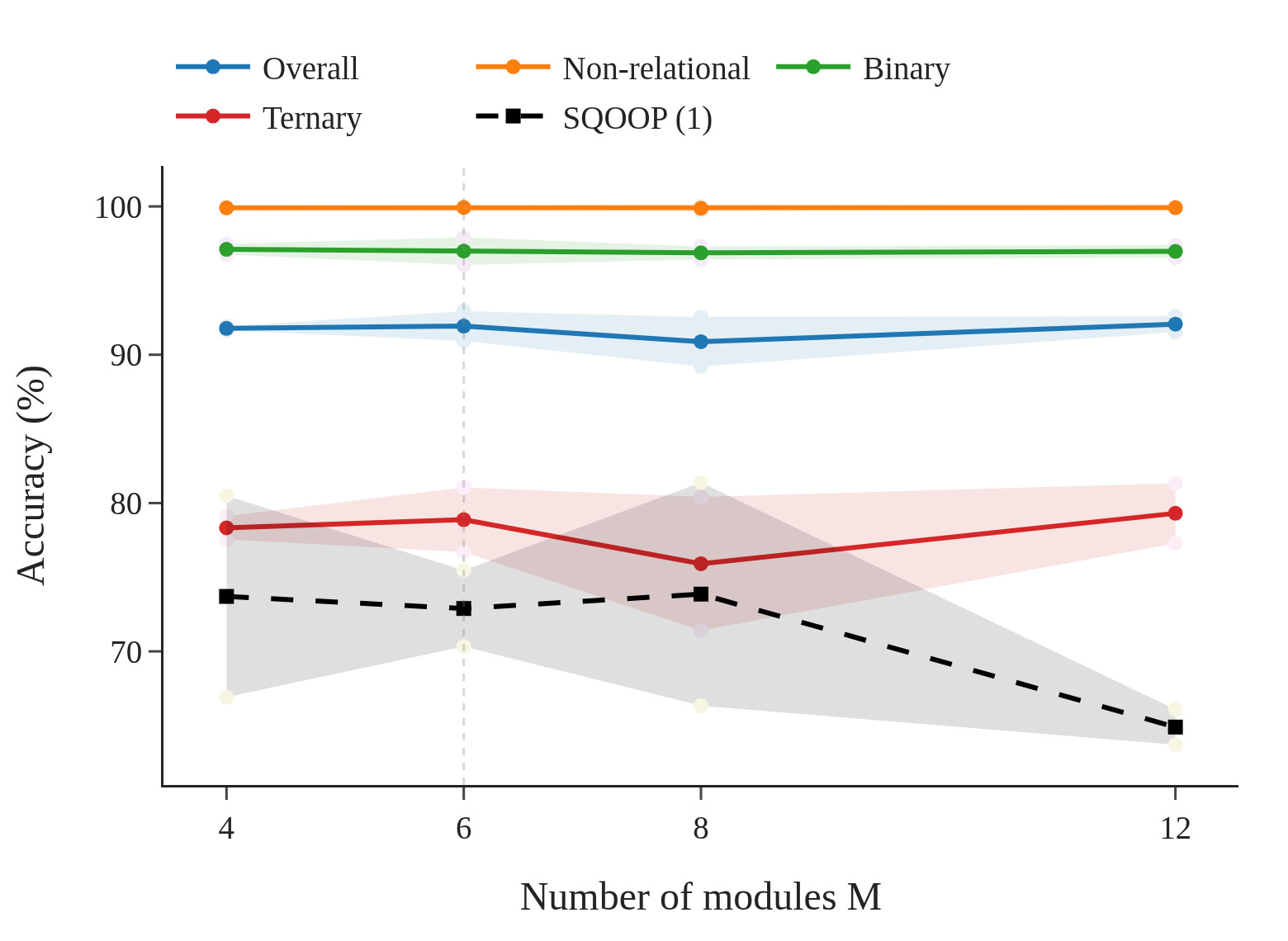}
    \caption{Number of modules $M$.}
    \label{fig:soc-ablations-modules}
\end{subfigure}
\caption{Sweeps around the main model, varying one setting with all else fixed; the dotted line marks the value used in the main model. Sort-of-CLEVR accuracy is shown by question family alongside SQOOP held-out accuracy at pair variety $1$.}
\label{fig:soc-ablations}
\end{figure}

These sweeps vary one setting at a time around the main model. The architecture makes one quantitative prediction, and it concerns the phase dimension. At most $d$ directions can be mutually orthogonal in $d$ dimensions, which bounds how many senders a receiver can separate at once. Accuracy should therefore rise with $d$ until there is enough capacity and flatten after that, and the gain should be largest where the most senders must be separated.

Figure~\ref{fig:soc-ablations}(a) shows this on Sort-of-CLEVR. Overall accuracy rises from $73.5\%$ at $d=2$ to $90.0\%$ at $d=3$, then sits between $91\%$ and $92\%$ through $d=8$. The ternary family gains most, from $56.3\%$ at $d=2$ to $73.7\%$ at $d=3$ and a peak of $79.1\%$ at $d=4$, and seed variation is widest at $d=2$. Two dimensions cannot hold the three orthogonal directions a three-way interaction needs, and the head must read from those three senders, giving four participants to separate. The peak at $d=4$ matches that count, though the argument bounds the directions available rather than showing the model uses them.

SQOOP does not follow the pattern. Held-out accuracy is highest at $d=2$ and then sits between $68.5\%$ and $72.9\%$, with wide variation throughout. Its questions name two objects rather than three, so the demand on capacity is lower and the flat curve is what the argument predicts. Why $d=2$ beats $d=3$ it does not explain, since the head participates here, too. We report the peak without an account of it.

Internal steps behave the other way round. Little changes beyond $T=2$ on Sort-of-CLEVR, while SQOOP rises from $57.8\%$ at $T=2$ to $73.0\%$ at $T=8$ before falling at $T=16$, so SQOOP benefits from more time for communication to develop. The number of modules has little effect up to $M=8$ on either task and degrades SQOOP thereafter, consistent with more senders sharing a phase space of fixed dimension.

\section{Ablation and override definitions}
\label{app:ablations}

\subsection{Ablations}

Each variant changes exactly one component of the main model and is trained from scratch under the protocol of Appendix~\ref{app:training}. All other components are those of Section~\ref{sec:method}.

\paragraph{No spatial preference.} The competition of Equation~\ref{eq:competition} assigns spatial cell $p$ to module $k$ by phase alignment with the module anchor, plus a learnt spatial preference $b_k(p)$ that encourages modules to claim distinct regions. The ablation removes it:
\begin{equation}
\text{no spatial preference:}\qquad b_k(p)=0,
\label{eq:abl-nobias}
\end{equation}
with $b_k(p)$ neither learnt nor used, leaving the assignment to emerge from phase alignment alone.

\paragraph{No structural modularity.} In the main model each module has its own recurrent cell $\operatorname{GRU}_k$ in Equation~\ref{eq:step-update}. The ablation replaces these with a single cell shared across modules:
\begin{equation}
\text{no structural modularity:}\qquad h^{(t)}_k = \operatorname{GRU}\bigl([X_k;r^{(t)}_k],h^{(t-1)}_k\bigr) \quad \text{for all } k,
\label{eq:abl-cells}
\end{equation}
communication is unchanged. The shared cell is not widened to compensate, so this variant has fewer parameters than the main model.

\paragraph{Communication.} The remaining ablations vary how communication is routed, combining an address source, phase or content, with a rule for combining messages, linear or softmax. The main model uses phase addresses with a linear read, Equation~\ref{eq:channel}, so called because the messages are combined with unnormalised alignment weights rather than through a softmax. The three alternatives complete the combinations:
\begin{align}
\text{phase + softmax:}\quad &r^{(t)}_k = \sum_{j\neq k}\operatorname*{softmax}_{j\neq k}\bigl(\tau\,\langle z^{(t-1)}_j,z^{(t-1)}_k\rangle\bigr)m^{(t)}_j, \label{eq:abl-softmax}\\
\text{content + linear:}\quad &z^{(t)}_k=\Pi(W_zh^{(t)}_k) \quad \text{at every step, replacing Equation~\ref{eq:phi}}, \label{eq:abl-content}\\
\text{content + softmax:}\quad &r^{(t)}_k=\sum_{j\neq k}\omega_{kj}m^{(t)}_j, \qquad \omega_{kj}=\operatorname*{softmax}_{j\neq k}\left(\tfrac{1}{\sqrt{d_h}}\langle W_Qh^{(t-1)}_k,W_Kh^{(t-1)}_j\rangle\right), \label{eq:abl-lines}
\end{align}
where $\tau$ is a softmax temperature. The first two keep the shared channel and change only one choice each. The third also replaces the shared channel with direct attention between modules, so its addresses are asymmetric and it changes both the address source and the communication topology; \textit{content + linear} is therefore the controlled comparison for the address source, as noted in Section~\ref{sec:results:ablations}.

\subsection{Overrides}

Overrides are applied to a trained model at test time. No parameter changes and no retraining take place, and each acts only on the module phases: content states $h^{(t)}_k$, module inputs $X_k$, the assignment weights $\alpha^{(t)}_{kp}$ and the anchors $\bar a_k$ are left as the trained model produces them. Perceptual modules are indexed $k=1,\ldots,M$ and the head $k=0$, as in Section~\ref{sec:method}. Each override is a parameter-free wrapper around the trained model, so the unmodified forward pass and the overridden one differ in exactly the lines below.

\paragraph{Freeze.} After the binding stage, set
\begin{equation}
z^{(t)}_k = z^{(0)}_k \quad \text{for } k = 1,\ldots,M \text{ and } t = 1,\ldots,T,
\end{equation}
discarding the output of $\Phi$ for those modules while retaining it for the head, whose phase evolves under Equation~\ref{eq:phi} as usual. The head is excluded deliberately: its initial phase is a random unit vector carrying no information about the scene, so a frozen head would read a fixed random projection of the channel at every step and the resulting collapse would measure a disabled readout rather than the perceptual modules' need for an evolving phase.

\paragraph{Zero.} Set $z^{(t)}_k=e$ for every $k$ including the head and every $t$, where $e$ is a single fixed unit vector. Every inner product in Equation~\ref{eq:channel} is then $1$ and each participant receives the unweighted average of the other messages. Addressing is removed; the messages are untouched.

\paragraph{Shuffle-phase.} Draw a permutation $\pi$ of $\{1,\ldots,M\}$ and, after the $T_b$ binding iterations of Equation~\ref{eq:competition} and before the first internal step, replace
\begin{equation}
z^{(0)}_k \leftarrow z^{(0)}_{\pi(k)} \quad \text{for } k = 1,\ldots,M,
\end{equation}
leaving $X_k$, $h^{(0)}_k$ and the head unchanged. Phases then evolve normally for all $T$ steps. The set of phases and the angles between them are preserved exactly; what changes is which content state each phase is attached to.

\paragraph{Shuffle-binding.} As above, but the permutation is applied to the bound pairs rather than the phases alone, so that $X_k$ and $z^{(0)}_k$ are permuted together and module $k$ receives module $\pi(k)$'s binding intact. The pairing of phase to content is preserved; what changes is which module holds which region.

\end{document}